\documentclass[conference]{IEEEtran}
\IEEEoverridecommandlockouts

\usepackage{multirow}
\usepackage{amsmath}
\usepackage{array}
\usepackage{microtype}
\usepackage{makecell}
\usepackage{xcolor}
\usepackage{listings}
\usepackage{longtable}
\usepackage{adjustbox}
\usepackage{caption}
\usepackage{booktabs}
\usepackage{graphicx}
\usepackage{tabularx}
\usepackage{fancyvrb}
\usepackage[T1]{fontenc}
\usepackage{tikz}
\usepackage{tcolorbox}
\usepackage{times}
\usepackage{latexsym}
\usepackage{enumitem}
\usepackage{graphicx}
\usepackage{subcaption}
\usepackage{pifont}

\usepackage{inconsolata}
\usetikzlibrary{shapes.geometric, arrows}
\definecolor{lightgray}{gray}{0.9}
\definecolor{electricblue}{rgb}{0.49, 0.98, 1}

\usepackage[colorlinks=true, urlcolor=blue, citecolor=black, linkcolor=black]{hyperref}
\usepackage{cite}
\usepackage{amsmath,amssymb,amsfonts}
\usepackage{algorithmic}
\usepackage{graphicx}
\usepackage{textcomp}
\usepackage{xcolor}
\def\BibTeX{{\rm B\kern-.05em{\sc i\kern-.025em b}\kern-.08em
    T\kern-.1667em\lower.7ex\hbox{E}\kern-.125emX}}

\newcommand{\stitle}[1]{\noindent\textup{\textbf{#1}}}

\begin{document}

\title{Persona-Guided LLM Agents for Task-Oriented Dialogue}

\author{
    \IEEEauthorblockN{
        Maryam Shoaeinaeini\IEEEauthorrefmark{1},
        Brent Harrison\IEEEauthorrefmark{1},
        A.B. Siddique\IEEEauthorrefmark{1}
    }
    \IEEEauthorblockA{
        \textit{University of Kentucky}\\
        Email: \{maryam.shoaei, brent.harrison, ab.siddique\}@uky.edu
    }
}
\maketitle

\begin{abstract}
Prior work has shown that large language models (LLMs) can express diverse personality traits in open-ended text generation. However, whether they can express these traits in goal-directed dialogue without compromising task performance, and whether personality adaptation improves user interactions, remain open questions. We investigate these questions in task-oriented dialogue (TOD) using a training-free framework in which a personality-conditioned user agent interacts with a system agent under three conditions: Neutral (no personality information), Try (adaptation to behavioral cues observed during dialogue), and Oracle (explicit access to the user's personality label). We evaluate these conditions across three LLMs (GPT-4o, Qwen3-Next-80B, and Gemini 2.0 Flash), two Schema-Guided Dialogue domains, and ten Big Five personality poles, comprising 9,000 dialogues evaluated by two independent LLM judges. User agents express most personality traits while the system agent maintains high task performance, achieving an overall completion rate of 89.94\% and a success rate of 85.27\%. \textit{Introversive} and \textit{Closed} are notable exceptions: these traits manifest primarily through behavioral absence and are rarely detected. Both adaptive conditions improve pairwise user satisfaction but reduce truthfulness, revealing a trade-off between personalization and grounding. However, only Try, which adapts to observed behavioral cues, also improves constraint satisfaction and inform rate. The effectiveness of adaptation depends on how personality information is used: Oracle's benefits vary with trait visibility, and explicit \textit{Disagreeable} labels lead Qwen3 and Gemini to mirror user hostility. Try avoids this failure, improves satisfaction regardless of trait visibility, and is preferred over Oracle overall. Finally, a human evaluation of 175 pairs provides additional evidence for the reliability of the LLM judges: in cases with a clear winner, annotators agree with the satisfaction judges more often than they agree with one another. These findings suggest that adapting to observable behavior is a more robust approach to social adaptation than conditioning on explicit personality labels. 
\end{abstract}



\section{Introduction}
\label{sec:introduction}

Large language models are increasingly used as assistants in settings where the quality of the interaction matters as much as the correctness of the final answer. Users want systems that not only complete tasks but also communicate in ways that feel appropriate, responsive, and personally aligned. Personalization has therefore become a central goal for dialogue agents, particularly in customer support, booking, and assistant scenarios, where interaction style can shape user trust and satisfaction \cite{park2023generative, wang2024rolellm}.

Prior work has shown that LLMs can express diverse personality traits in open-ended text generation. They imitate recognizable personality patterns in questionnaires, narratives, role-play, and controlled generation \cite{li2022gpt,pan2023llms,serapio2023personality,jiang2023evaluating,jiang2024personallm,mao2023editing}, exhibit measurable Big Five traits \cite{goldberg2013alternative}, and maintain persona-consistent patterns in generated text. However, most evaluations study personality expression outside goal-directed dialogue. It therefore remains unclear whether LLMs can express personality while pursuing a concrete task, and whether adapting to the user's personality actually improves the interaction.

This question is especially important in task-oriented dialogue (TOD), where a system helps a user accomplish a goal through multi-turn interaction. Unlike open-ended conversation, TOD systems must satisfy schema constraints, gather required slot values, issue valid API calls, and complete the user's task \cite{rastogi2020towards,zang2020multiwoz,mosharrof2025evaluating}. Personality-aware behavior may improve the interaction by making the system more responsive to the user's communication style, uncertainty, or preferences. It may also introduce risks: adapting too strongly to personality cues could weaken grounding, distort recommendations, or interfere with task completion. Personality-aware TOD thus requires balancing personalization against task-grounded behavior.

Motivated by this gap, we study three research questions.
\textbf{RQ1}: In a two-agent TOD interaction, can the user agent realize a target personality trait while the system maintains schema-guided task performance?
\textbf{RQ2}: How does access to the user's personality affect system quality and pairwise user satisfaction?
\textbf{RQ3}: How does the strength of user trait realization shape the satisfaction gains from personality conditioning, and does this relationship differ between explicit (\textit{Oracle}) and implicit (\textit{Try}) conditioning?

\begin{figure*}[t!]
  \centering
  \includegraphics[width=0.9\textwidth]{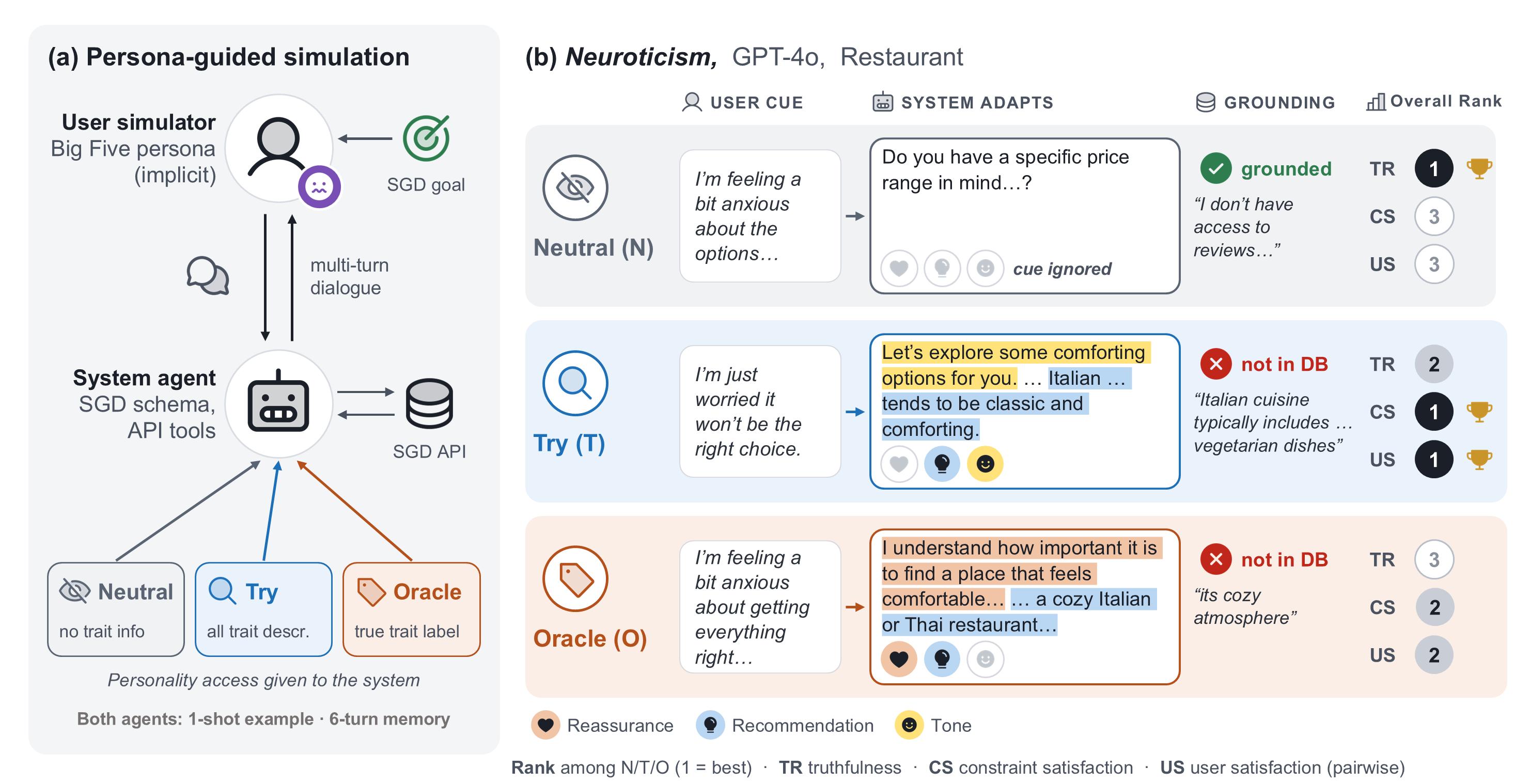}

\caption{Overview of the framework. (a) A user simulator with an implicit Big Five persona talks to an API-grounded system agent that receives no user's trait information (Neutral), all trait descriptions (Try), or the true trait label (Oracle). (b) For a \textit{Neuroticism} user, adaptation adds reassurance and better-tailored recommendations but lowers grounding, while Try achieves the highest user satisfaction. Ranks by mean score for Neuroticism users, averaged over models and domains}

  \label{fig:overview}
\end{figure*}

To answer these questions, we build a training-free framework for testing how personality adaptation affects LLM agents in schema-guided TOD. We compare three system conditions: \textit{Neutral}, where the system receives no personality information; \textit{Try}, where it is asked to adapt to the user's personality from dialogue cues alone; and \textit{Oracle}, where it is given the user's target trait explicitly. This design isolates the effect of personality access while keeping domain structure, schemas, and task requirements fixed. We evaluate GPT-4o, Qwen3-Next-80B, and Gemini 2.0 Flash on Hotel and Restaurant dialogues from the Schema-Guided Dialogue (SGD) dataset \cite{rastogi2020towards}, covering the Big Five traits and their opposite poles (9,000 dialogues). We assess the dialogues along five dimensions: task outcomes, personality realization, system quality, system adaptation, and pairwise user satisfaction. To reduce dependence on a single evaluator, every judge-based metric is scored by two independent LLM judges (Gemma-4-31B and GPT-5-mini) and averaged. Fig.~\ref{fig:overview} summarizes the framework of persona-guided dual agents along with an example dialogues for each condition. 

Our findings answer the three questions as follows. For \textbf{RQ1}, user agents express most traits while the system maintains strong task performance (89.94\% completion, 85.27\% success); only \textit{Introversive} and \textit{Closed} are rarely detected, consistent with traits defined by absent behavior being hard to observe in short task dialogue. For \textbf{RQ2}, personality access makes users more satisfied: averaged over two judges, both Oracle (57\% vs.\ 39\%) and Try (60\% vs.\ 35\%) are preferred over Neutral, and Try is slightly preferred over Oracle (52\% vs.\ 43\%). Only Try significantly improves constraint satisfaction and inform rate, while both adaptive conditions reduce truthfulness, revealing a tension between personalization and task grounding. Task-success gains appear only in the preference-rich Restaurant domain. For \textbf{RQ3}, the two strategies diverge. Oracle's satisfaction gains scale with how visibly the user expresses the trait (Spearman $\rho{=}0.69$), whereas Try's do not ($\rho{=}0.13$). Oracle falls below Neutral when the trait is barely visible (\textit{Introversive}, \textit{Closed}), and for \textit{Disagreeable} users it leads Qwen3 and Gemini to mirror hostility (Oracle loses to Neutral 32\% vs.\ 68\%), while Try still beats Neutral in all three cases. A condition-blind System Adaptation Score confirms that both strategies change system behavior (Neutral 2.22, Try 2.70, Oracle 3.40), mostly for expressive traits, and that Oracle does not make the user's trait itself more visible. A 175-pair human study supports both satisfaction judges: annotators agree with the judges more often than with each other, both including ties (59.0\% vs.\ 47.4\%; $\kappa{=}0.337$ vs.\ $0.189$) and on clear-winner cases (76.3\% vs.\ 65.1\%; $\kappa{=}0.525$ vs.\ $0.304$), rising to 83.2\% ($\kappa{=}0.663$) when the two judges agree with each other.

Together, these results support a design principle for social agents: adapting to observable behavior is more robust than conditioning on an explicit psychological label.\footnote{Source code: \url{https://github.com/CORGI-lab/personality\_TOG}}

\section{Related Work} 

\stitle{LLM-based Task-Oriented Dialogue.}
Task-oriented dialogue (TOD) research has traditionally focused on systems that track dialogue state, fill slots, predict system actions, and ground responses in structured APIs or knowledge sources. Progress in this area has depended heavily on high-quality annotated datasets such as MultiWOZ and Schema-Guided Dialogue (SGD), which support benchmarking across domains but also highlight the cost and brittleness of manual annotation \cite{eric2020multiwoz,zang2020multiwoz,han2021multiwoz,rastogi2020towards}. 
Benchmarks such as schema adherence, state tracking, and completion of user goals, which formalize the core requirements of TOD systems, remain central even when the generator is an LLM. Recent work has increasingly replaced fully supervised pipelines with LLM-based approaches that use in-context learning or reduced annotation to generalize to new domains more efficiently \cite{mosharrof2025evaluating}. This shift improves flexibility, but the primary focus remains task success, slot accuracy, and robustness under domain shift rather than user-specific adaptation. As a result, prior TOD work provides strong foundations for structured interaction, but leaves open how personality cues alter task completion, grounding, and user satisfaction in schema-constrained settings.

\stitle{Personality and Persona Control in LLMs.}
A separate line of work studies whether large language models can express stable personality traits or personas. Prior studies evaluate personality signals through questionnaire-style probes, narrative generation, and controlled prompting, showing that LLMs can imitate recognizable trait patterns and exhibit measurable variation across personality dimensions \cite{li2022gpt,pan2023llms,serapio2023personality,jiang2023evaluating,jiang2024personallm}. Related work on role-playing and agent simulation further suggests that LLMs can maintain social roles and behavioral framing over longer interactions \cite{wang2024rolellm,park2023generative}. Other studies move from measuring personality to actively inducing or editing it, for example through prompt design, persona conditioning, or model-level control mechanisms \cite{mao2023editing,lotfi2023personalitychat}. However, these studies mostly evaluate personality realization in open-ended or non-task settings. They therefore show that personality can be represented and manipulated in LLM outputs, but do not establish how personality conditioning interacts with schema constraints, API decisions, or multi-turn task completion.

\stitle{Personalized Dialogue and User Modeling.}
Personalized dialogue research studies how systems adapt to user characteristics, preferences, or inferred conversational tendencies. Recent surveys frame this direction as part of LLM personalization, where models condition on user-specific context, such as facts, style, persona descriptions, or trait-like profiles, to make conversations more relevant and coherent over time \cite{tseng2024two,zhang2018personalizing,lotfi2023personalitychat}. Recent LLM-based work has also explored persona-aware dialogue generation, where user profiles or dialogue history are used to improve persona consistency across interactions \cite{liu2025persona}. This line of work is related to our focus on personality-aware interaction, but it is not designed for schema-guided task-oriented dialogue, where systems must satisfy slot constraints, issue valid API calls, and complete user goals. In parallel, LLM-based TOD user simulation uses prompted LLMs to generate goal-driven user behavior for training or evaluating dialogue systems \cite{algherairy2025prompting,davidson2023user}.

\section{Proposed Framework}
\label{sec:realtod}

We build a controlled framework for evaluating personality-aware task-oriented dialogue on the Schema-Guided Dialogue (SGD) dataset \cite{rastogi2020towards}. An LLM-based user agent interacts with an LLM-based system agent under matched task settings. The task schema, domain, demonstration dialogue, and generation rules are held fixed across conditions; the only experimental variable is the system agent's access to personality information (\textit{Neutral}, \textit{Try}, or \textit{Oracle}; Section~\ref{sec:introduction}).

\subsection{Task-Oriented Dialogue Framework}

Each dialogue is grounded in an SGD task schema that specifies the domain, intents, slots, and API structure required to complete the task. Both agents are prompted one-shot, without fine-tuning: at each turn $t$, generation is conditioned on an in-domain demonstration dialogue, the schema and API rules, operational instructions, and the dialogue history up to $t$.

At each turn, the system agent produces either a natural-language response or an API call. Following \cite{mosharrof2025evaluating}, we represent an API call as
\begin{equation}
\text{APICall} = \big(I, \{(s_i, v_i)\}\big),
\end{equation}
where $I$ is the domain intent and $(s_i, v_i)$ are slot--value pairs. This representation lets us check whether the dialogue stays grounded in the schema and whether the system completes the search or reservation with valid slot values.

\subsection{Personality Trait Specification}

We model user personality with the Big Five framework, which captures broad behavioral variation along five dimensions \cite{digman1990personality,goldberg2013alternative,wiggins1996five,de2000big,jiang2024personallm}. To cover both poles of each dimension, we use ten target traits: \textit{Extraversion} and \textit{Introversive}, \textit{Agreeableness} and \textit{Disagreeable}, \textit{Conscientiousness} and \textit{Unconscientious}, \textit{Neuroticism} and \textit{Stable}, and \textit{Openness} and \textit{Closed}. Trait descriptions follow the Personality Prompting (P\textsuperscript{2}) descriptions of \cite{jiang2023evaluating}, for both the five canonical traits and their opposite poles.

Each user agent is assigned exactly one target trait and is instructed to express it implicitly through its behavior, without naming it. This lets us test whether personality can be realized in task-oriented interaction while the system still satisfies schema and API constraints.

\subsection{Prompting and Personality-Access Conditions}

Both agents share the same prompt structure: task instructions, domain schema and API specification, an in-domain demonstration dialogue, operational rules, and the dialogue history. Personality information is added only where the condition requires it.

The user agent receives the target trait and its description, which guide how the user asks questions, reacts to recommendations, handles uncertainty, and decides whether to book.

The system agent is always instructed to remain task-valid, grounded, and schema-compliant: it issues API calls only after the required slots are collected, uses only schema fields and search results, and confirms key slots before finalizing a reservation. The three conditions differ only in personality access:
\begin{itemize}
    \item \textit{Neutral}: the system receives no personality information.
    \item \textit{Try}: the system is instructed to adapt to the user's personality as revealed in the dialogue, without receiving a label.
    \item \textit{Oracle}: the system receives the user's target trait and its description.
\end{itemize}
In both personality-aware conditions, the system may adjust its level of detail, reassurance, recommendation style, or clarification strategy. We do not test whether Try recovers the latent trait, so we interpret it as \emph{cue-responsive} adaptation: the system responds to observable behavior such as verbosity, hesitation, or hostility.

\section{Experiments}




\subsection{Datasets}
We use the Restaurant and Hotel domains of the Schema-Guided Dialogue (SGD) dataset \cite{rastogi2020towards}. Both involve schema-guided search and reservation, but they differ on the axis most relevant to personality adaptation. Restaurant is preference-driven (cuisine, price range, ambiance), giving the system latitude to tailor recommendations to the user, whereas Hotel is constraint-driven (location, star rating, rooms), leaving less room to adapt. This contrast lets us examine whether the benefits of personality access depend on how much latitude the task provides. Before generation, we remove duplicate and inconsistent service records from both databases without altering the task schemas.

\subsection{Experimental Setup}
We use the same experimental structure for GPT-4o, Qwen3-Next-80B, and Gemini 2.0 Flash, with only minor model-specific prompt formatting to ensure valid API calls. For each combination of domain, model, access condition, and target trait, we generate 50 dialogues, yielding 9,000 dialogues in total. The user always initiates the interaction, and a dialogue ends when the user produces the scripted closing phrase after completing the task, or when it reaches 70 turns or 5 API calls. Dialogue generation uses temperature 0.7, top-$p$ 1.0, and at most 250 output tokens per call. GPT-4o is accessed through OpenAI, Gemini 2.0 Flash and GPT-5-mini through OpenRouter, and Qwen3-Next-80B and Gemma 4 through Ollama Cloud.

\subsection{Evaluation Metrics}
We evaluate the dialogues with automatic and LLM-judged metrics. To reduce dependence on a single evaluator, every judged metric is scored independently by two LLM judges, Gemma 4 and GPT-5-mini, and we report the average of the two.

\begin{enumerate}
    \item \textbf{Task outcomes.} Both task outcomes are computed automatically from the dialogue, without an LLM judge. \textit{Dialogue Completion} equals $1$ if the dialogue ends with the scripted user phrase \emph{``Thanks a lot! That's all I need''} before reaching the 70-turn or 5-API-call limit, and $0$ otherwise. \textit{Task Success} equals $1$ if the final reservation \texttt{APICall} is issued with the correct required slot values, and $0$ otherwise. The two can diverge: a dialogue may end naturally without the user ever requesting a reservation (Completion $=1$, Success $=0$), and a dialogue that exhausts its turn or API-call budget has Completion $=0$.

    \item \textbf{Personality realization.} The \emph{User Trait Score} measures how visibly the user expresses the target trait. The judge reads the dialogue, identifies the three most evident user traits among the ten candidates, and rates each on $1$--$5$. We report the score of the \emph{intended} trait, or $0$ if it is not among the top three, giving an effective range of $0$--$5$. Near-zero values mean the trait is rarely among the most visible behaviors. We separately verify that each trait prompt points in the intended Big Five direction with a 44-item BFI probe \cite{john1999big}, answered on a 1--5 Likert scale from the trait description rather than from generated dialogues.

    \item \textbf{System quality.} The judge scores three dimensions on $1$--$5$ (higher is better). \emph{Constraint Satisfaction} measures whether the final recommendation or booking fulfills the user's active constraints. \emph{Inform Rate} measures whether the system provides the requested information correctly and as supported by the dialogue and search results. \emph{Truthfulness} measures whether the system avoids unsupported or fabricated claims. The judge is given the fields available in each domain's database, and any specific claim beyond these fields that the dialogue does not support counts as a hallucination.

    \item \textbf{Pairwise user satisfaction.} We pair dialogues matched on domain, model, and target trait but differing in access condition (3,000 pairs per comparison), and the judge selects which system the user would likely find more satisfying, or a tie. We report win rates for Oracle vs.\ Neutral, Try vs.\ Neutral, and Try vs.\ Oracle. The judge is instructed to ignore task success (measured separately), not to penalize empty search results unless the system handles them poorly, not to favor shorter dialogues, and not to treat the scripted closing phrase as evidence of satisfaction. To control for position bias, both judges see each pair in a randomly assigned A/B order. Because no real user with the target personality rates the interaction, this is an observer judgment of presumed satisfaction, which we validate against human annotators in \S\ref{sec:human}.

    \item \textbf{System adaptation.} The \emph{System Adaptation Score} ($1$--$5$) measures how well the system adapts its tone, pacing, reassurance, and recommendation style to the user's target trait. The judge sees the dialogue and the target trait but not the access condition that produced it, so its score reflects observable system behavior rather than the presence of personality information in the system prompt.

\end{enumerate}

\section{Results and Analysis}
\label{sec:results}

We answer the three research questions from Section~\ref{sec:introduction}
by comparing Neutral, Try, and Oracle on task performance, personality
realization, system quality, and pairwise satisfaction in both domains.
Statistical tests are reported within each RQ, and §\ref{sec:human}
validates the satisfaction judges against human annotators.
\subsection{\textbf{RQ1}: Trait Realization and Task Performance}
\label{sec:rq1_results}

Table~\ref{tab:rq1_trait_feasibility} summarizes task and personality
outcomes averaged across domains, models, and system conditions. Task
performance is strong: 89.94\% Dialogue Completion and 85.27\% Task
Success. Personality conditioning does not generally prevent dialogues
from reaching a natural endpoint or completing the reservation task.
 
Personality expression is clear for most traits. The average is 3.520, with strong realization for \textit{Extraversion}, \textit{Agreeableness}, \textit{Neuroticism}, \textit{Openness}, and \textit{Disagreeable}. \textit{Introversive} and \textit{Closed} are the clear exceptions, with near-zero
scores (1.042 and 0.078). These low scores are consistent with a behavioral-signal problem. A neutral user and system control establishes why: 100 Restaurant dialogues with an unconditioned user show that the Gemma 4 judge default conversational persona is judged as \textit{Stable} (74\%) and \textit{Conscientious} (26\%). Judge-free counts confirm that users with these traits give fewer personality cues, so judges read them as the model’s default persona.
A one-sided binomial test against the 30\% chance baseline (the probability of any label appearing in the top-3 under uniform selection from 10 candidates) quantifies this: eight of ten traits are detected significantly above chance after Holm--Bonferroni correction (all $p_{\text{holm}} \approx 0$), while \textit{Introversive} at 28.3\% ($p_{\text{holm}} = 1.00$) is indistinguishable from random guessing and closed at 3.1\% falls below it, displaced by \textit{Stable} and \textit{Agreeableness}. A judge-free count over all 9,000 dialogues confirms this: \textit{Introversive} and closed average 7.83/7.03 and 7.13/6.90 words/turn (Hotel/Restaurant) vs. 13.4–36.8/11.9–33.5 for other traits.

The BFI probe yields a high average score of 4.888, confirming that
the trait prompts align well with their intended Big Five directions.
The gap between strong prompt-level alignment and uneven
dialogue-level expression shows that some traits are simply harder to
realize in short, goal-directed turns than in open-ended generation.
Overall, personality-aware TOD is feasible for most traits.

\begin{table}[t!]
\centering
\caption{Trait-level task and personality realization outcomes averaged across domains, models, and conditions. User trait score is average over two llm judges responses.}
\resizebox{\columnwidth}{!}{%
\begin{tabular}{l|cc|cc}
\toprule
\textbf{Trait} &
\multicolumn{2}{c|}{\textbf{Task Outcomes}} &
\multicolumn{2}{c}{\textbf{Personality Realization}} \\
\cmidrule(lr){2-3} \cmidrule(lr){4-5}
& \makecell{\textbf{Dialogue}\\\textbf{Completion (\%)}} &
  \makecell{\textbf{Task}\\\textbf{Success (\%)}} &
  \makecell{\textbf{User Trait}\\\textbf{Score}} &
  \makecell{\textbf{BFI}\\\textbf{Score}} \\
\midrule
\textit{Extraversion}      & 79.67 & 82.12 & 4.918 & 5.000 \\
\textit{Agreeableness}     & 95.89 & 87.44 & 4.843 & 4.963 \\
\textit{Conscientiousness} & 93.89 & 83.34 & 4.431 & 5.000 \\
\textit{Neuroticism}       & 77.22 & 75.66 & 4.938 & 4.967 \\
\textit{Openness}          & 85.33 & 81.12 & 4.847 & 4.967 \\
\midrule
\textit{Introversive}      & 96.34 & 91.34 & 1.042 & 4.625 \\
\textit{Disagreeable}      & 78.78 & 76.22 & 4.648 & 5.000 \\
\textit{Unconscientious}   & 96.34 & 95.22 & 2.343 & 4.778 \\
\textit{Stable}            & 97.89 & 88.22 & 3.109 & 4.917 \\
\textit{Closed}            & 98.11 & 92.00 & 0.078 & 4.667 \\
\midrule
\textbf{Average}  & \textbf{89.94} & \textbf{85.27} & \textbf{3.520} & \textbf{4.888} \\
\bottomrule
\end{tabular}%
}
\label{tab:rq1_trait_feasibility}
\end{table}

\subsection{RQ2: Effect of Personality Conditioning on System Quality and User Satisfaction}
\label{sec:rq2}
Table~\ref{tab:rq2_system_quality_satisfaction} compares system-quality metrics and pairwise satisfaction preferences from two LLM judges across traits, averaged over domains and LLMs. Try achieves the highest average Constraint Satisfaction (4.530) and Inform Rate (4.720), followed by Oracle (4.489 and 4.694), while Neutral is lowest on both (4.409 and 4.638). Because conditions are simulated independently, we test differences with Wilcoxon signed-rank tests over the 60 matched domain $\times$ LLM $\times$ trait cells (Holm-corrected). Try significantly improves Constraint Satisfaction ($p = .002$) and Inform Rate ($p = .004$) over Neutral, whereas Oracle's gains are not significant. Both adaptive conditions significantly reduce Truthfulness (Try $p = .001$; Oracle $p < .001$), a drop that holds under each judge separately, revealing a tension between personalization and task grounding.

Personality-aware conditions are also preferred in pairwise comparisons. Averaged over two judges, Oracle outperforms Neutral 57\% to 39\% and Try leads Neutral 60\% to 35\% meaning that personality aware systems can make users more satisfied. Comparing the two adaptive strategies directly, Try is slightly preferred over Oracle overall (52\% vs.\ 43\%). Oracle performs well against Neutral for \textit{Openness} (82\% vs.\ 17\%), \textit{Extraversion} (73\% vs.\ 26\%), and \textit{Agreeableness} (70\% vs.\ 25\%), where the target style is socially clear and easy to support. Try is preferred over Oracle for \textit{Introversive}, \textit{Disagreeable}, \textit{Closed}, \textit{Unconscientious}, and \textit{Neuroticism}. The \textit{Disagreeable} trait is the sharpest example: Oracle underperforms Neutral (32\% vs.\ 68\%), while Try remains competitive (55\% vs.\ 44\%) and is strongly preferred over Oracle (77\% vs.\ 23\%). Under both judges, this collapse under Oracle is driven by Qwen3 and Gemini (O/N between 10/90 and 28/72 across domains), which mirror the hostile user tone rather than adapting constructively, a failure less pronounced under Try, which responds to observed behavior rather than committing to an explicit label. GPT-4o does not exhibit this failure (O/N 42/55 in Hotel and 71/28 in Restaurant). Human raters agree with the judge's preferences on clear-winner cases (\S\ref{sec:human}).

Domain characteristics modulate these gains. Personality access improves Task Success in Restaurant (Neutral$\rightarrow$Oracle: 78.07\%$\rightarrow$82.80\%, with gains for both GPT-4o and Gemini) but slightly reduces it in Hotel (91.53\%$\rightarrow$88.87\%). We hypothesize that this reflects preference latitude: Restaurant choices depend on cuisine, price, and ambiance, giving the system room to adapt, whereas Hotel choices are constrained by location, rating, and rooms. With two domains, however, preference latitude is confounded with other domain differences, and testing it requires domains that vary along this axis.

Taken together, Try provides the strongest overall balance: the highest Constraint Satisfaction, Inform Rate, and user satisfaction over Oracle and Neutral, and no tone-mirroring failure of the kind Oracle triggers for disagreeable users in Qwen3 and Gemini. Gains are largest in Restaurant domain which is a preference-rich domain.

\begin{table*}[t]
\centering
\caption{Trait-level system-quality scores and pairwise
satisfaction preferences averaged across domains and LLMs.
N, T, O: Neutral, Try, Oracle. CS, IR, TR, US: Constraint
Satisfaction, Inform Rate, Truthfulness (1--5, higher is better), User Satisfaction (pairwise preference dialogue).
All metrics are averaged over two judges (Gemma 4 and GPT-5-mini).
Pairwise cells report win-rate percentages for the first condition
over the second; values may sum to less than 100 due to ties.}
\small
\begin{tabular}{l|ccc|c}
\toprule
\textbf{Trait}
& \textbf{CS (N/T/O)}
& \textbf{IR (N/T/O)}
& \textbf{TR (N/T/O)}
& \textbf{US(O/N, T/N, T/O)} \\
\midrule
\textit{Extraversion}      & 4.198/4.240/4.175 & 4.418/4.452/4.352 & 4.515/4.203/3.930 & 73/26, 60/36, 40/57 \\
\textit{Agreeableness}     & 4.645/4.783/4.743 & 4.787/4.893/4.853 & 4.832/4.840/4.775 & 70/25, 62/31, 36/57 \\
\textit{Conscientiousness} & 4.555/4.642/4.672 & 4.680/4.775/4.800 & 4.852/4.777/4.780 & 62/34, 53/41, 40/53 \\
\textit{Neuroticism}       & 3.968/4.090/3.828 & 4.498/4.548/4.492 & 4.540/4.273/4.222 & 64/35, 68/31, 59/40 \\
\textit{Openness}          & 4.200/4.272/4.340 & 4.580/4.593/4.558 & 4.557/4.328/3.948 & 82/17, 59/39, 29/69 \\
\midrule
\textit{Introversive}      & 4.713/4.842/4.753 & 4.838/4.910/4.878 & 4.883/4.855/4.790 & 38/55, 62/30, 68/26 \\
\textit{Disagreeable}      & 3.695/3.953/3.938 & 4.193/4.395/4.402 & 4.545/4.372/4.512 & 32/68, 55/44, 77/23 \\
\textit{Unconscientious}   & 4.603/4.800/4.775 & 4.672/4.827/4.815 & 4.740/4.772/4.767 & 54/43, 62/33, 58/35 \\
\textit{Stable}            & 4.685/4.830/4.808 & 4.835/4.887/4.878 & 4.833/4.813/4.807 & 52/39, 59/31, 48/42 \\
\textit{Closed}            & 4.830/4.852/4.860 & 4.882/4.917/4.910 & 4.827/4.847/4.867 & 42/49, 55/34, 63/28 \\
\midrule
\textbf{Average}  & \textbf{4.409/4.530/4.489} & \textbf{4.638/4.720/4.694} & \textbf{4.712/4.608/4.540} & \textbf{57/39, 60/35, 52/43} \\
\bottomrule
\end{tabular}
\label{tab:rq2_system_quality_satisfaction}
\end{table*}

\subsection{RQ3: Trait Realization and User Satisfaction Gain under Explicit (\textit{Oracle}) and Implicit (\textit{Try}) Personality Conditioning}
\label{sec:rq3}

\begin{figure}[t]
  \centering
  \includegraphics[width=\columnwidth]{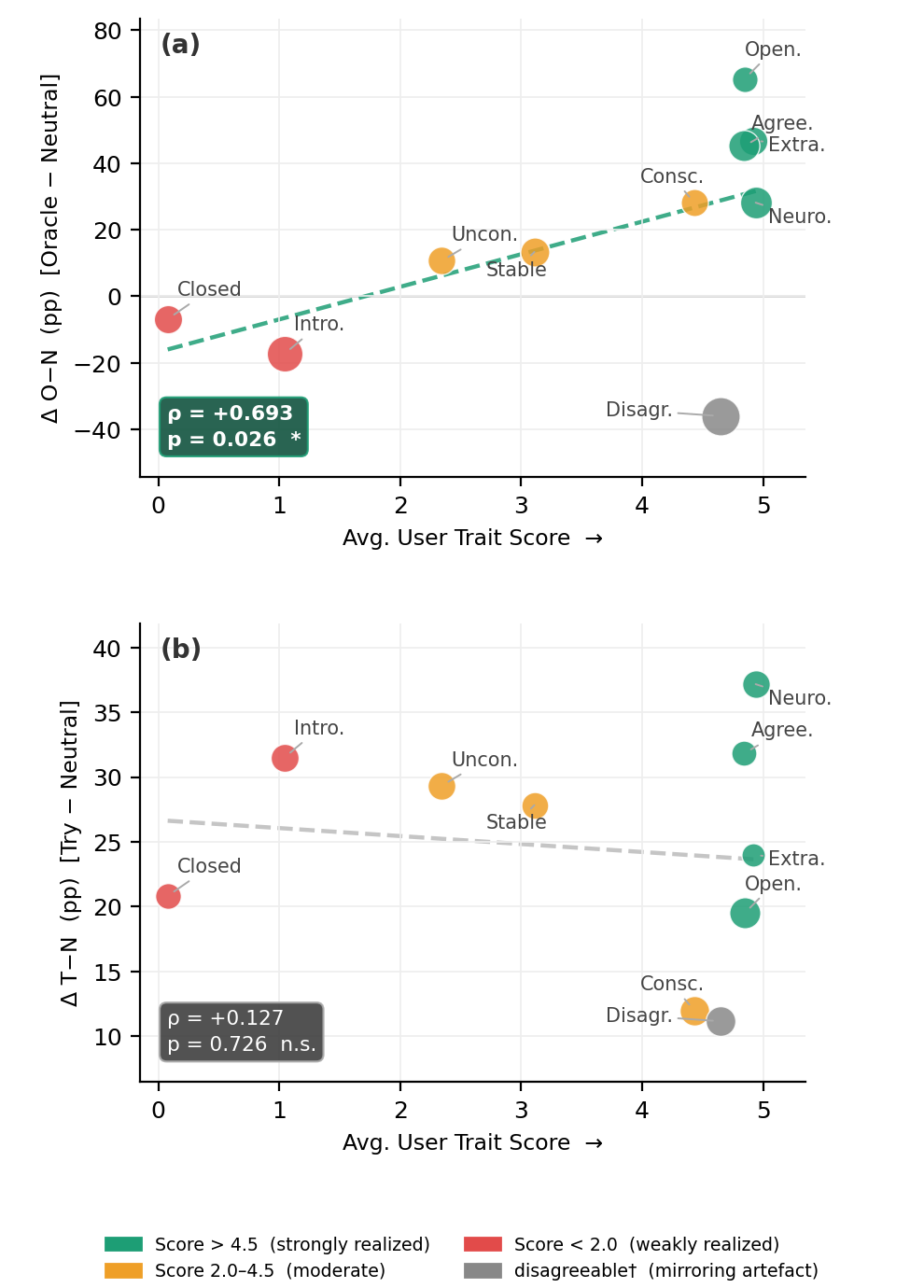}
\caption{Spearman correlation between Avg.\ User Trait Score and pairwise satisfaction deltas ($n{=}10$). The gray dashed line is a Theil–Sen trend line, and point size shows the SD across the six model–domain cells.}
  \label{fig:realization_main}
  \vspace{-0.0em}
\end{figure}

We correlate User Trait Score with pairwise satisfaction gains using Spearman rank correlation (Fig.~\ref{fig:realization_main}), which suits the bounded, non-normal score distribution (six traits score 4.4--4.9, \textit{Stable} and \textit{Unconscientious} are intermediate at 3.1 and 2.3, and \textit{Introversive} and \textit{Closed} are near zero at 1.0 and 0.1). Both User Trait Score and pairwise preferences are averaged over two judges (Gemma 4 and GPT-5-mini). We define $\Delta_{O\text{-}N}$ as the Oracle win rate minus the Neutral win rate (pp) over the 50 pairwise comparisons in each model$\times$domain$\times$trait cell, and $\Delta_{T\text{-}N}$ analogously; trait-level values average the six model--domain cells.

Oracle gains scale strongly with trait realization: $\Delta_{O\text{-}N}$ correlates with User Trait Score at $\rho = +0.693$ ($p = .026$, $n = 10$). This holds at the model$\times$domain$\times$trait cell level as well ($\rho = +0.479$, $p < .001$, $n = 60$), and within models it is significant for GPT-4o ($\rho = +0.621$, $p = .004$) and Gemini ($\rho = +0.569$, $p = .009$) but not Qwen3 ($\rho = +0.363$, $p = .115$). Try gains show no such dependence: $\Delta_{T\text{-}N}$ is not significantly correlated with User Trait Score at either the trait level ($\rho = +0.127$, $p = .726$) or the cell level ($\rho = +0.085$, $p = .519$). The divergence has a clear explanation. Oracle conditions on the trait \emph{label}, so its benefit requires the label to match observable user behavior. When realization is weak --- as for \textit{Introversive} and \textit{Closed} --- the label provides little evidence the system can act on, and Oracle falls below Neutral (\textit{Introversive}: 38/55; \textit{Closed}: 42/49 in Table~\ref{tab:rq2_system_quality_satisfaction}). Try adapts to \emph{dialogue cues} rather than a predefined label, so it remains useful even when the intended trait is not strongly expressed, still beating Neutral for \textit{Introversive} (62/30) and \textit{Closed} (55/34). The same mechanism explains the \textit{Disagreeable} Oracle collapse: Qwen3 and Gemini mirror the hostile label rather than adapting to the actual dialogue, so high realization coincides with negative satisfaction. Excluding the four \textit{Disagreeable} cells for Qwen3 and Gemini (two models $\times$ two domains) raises the cell-level correlation to $\rho = +0.633$ ($p < .001$, $n = 56$), indicating that mirroring is a model-specific prompt-following failure rather than a general property of personality conditioning.

A condition-blind System Adaptation Score independently confirms this chain. 
To measure how far the system adapts its behavior (tone, pacing, reassurance, directness, and recommendation style) to the user's personality, two LLM judges (Gemma 4 and GPT-5-mini) rated each dialogue on a 1--5 \emph{System Adaptation Score}. Judges were given the intended user trait and its definition, and assessed only the system's turns: whether its tone, pacing, reassurance, directness, flexibility and recommendation strategy fit the user's trait. They were instructed not to reward politeness or task completion on their own. Table~\ref{tab:sys_adaptation} reports the scores averaged over both judges. Adaptation increases steadily from Neutral (2.22) to Try (2.70) to Oracle (3.40). Both gains are significant (Wilcoxon signed-rank over 60 matched cells, Holm-corrected: T$-$N $+0.48$, O$-$N $+1.18$, O$-$T $+0.71$, all $p<.001$), and the same ordering holds for each judge separately. Inferring the trait from the conversation alone (Try) therefore recovers about 40\% of the gain obtained when the trait is given explicitly (Oracle).
The gains are uneven across traits. Adaptation is strongest for traits with salient conversational markers: \textit{Extraversion} ($+2.72$), \textit{Openness} ($+2.36$), \textit{Disagreeable} ($+1.88$) and \textit{Neuroticism} ($+1.52$). Under Try, \textit{Extraversion} and \textit{Neuroticism} already reach most of their Oracle-level adaptation. In contrast, systems barely adapt to low-arousal or low-expressivity traits: \textit{Introverted}, \textit{Stable} and \textit{Closed} improve by at most $0.35$ even when the trait is given. This suggests that current LLM systems adapt mainly by mirroring overt affect and energy, and rarely adjust to quieter or more reserved users.
The two judges differ in absolute level. Gemma 4 assigns the minimum score whenever adaptation is not explicit (Neutral average 1.38), whereas GPT-5-mini already rewards a calm, organized system under Neutral (3.06), especially for \textit{Stable}, \textit{Conscientiousness} and \textit{Introverted} users. As a result, raw cell scores correlate only weakly across judges (Spearman $\rho=0.15$). However, the judges agree closely on \emph{which} traits benefit from adaptation: across the 10 traits, $\rho=0.92$ for the O$-$N gain and $\rho=0.88$ for the T$-$N gain. We therefore interpret the condition differences, rather than absolute score levels.

A Friedman test rules out the alternative explanation that Oracle gains
simply reflect improved trait detectability. User Trait Score, averaged
over the two judges, does not differ significantly across conditions at
the trait$\times$LLM$\times$domain cell level ($n=60$; $\chi^2 = 1.14$,
$df = 2$, $p = 0.566$; all pairwise post-hoc $p \geq 0.56$ after Holm
correction; means: Neutral $= 3.53$, Try $= 3.50$, Oracle $= 3.52$).
The result holds for each judge separately (Gemma 4: $\chi^2 = 0.21$,
$p = 0.902$; GPT-5-mini: $\chi^2 = 2.85$, $p = 0.240$). Oracle does not
make the user's trait more visible --- it only helps when the trait was
already visible. This is the \emph{label--behavior match effect}:
explicit conditioning works best when user behavior already provides
evidence for the supplied label. Try's satisfaction gains being insensitive to realization strength is consistent with responding to observable cues rather than
to the trait label itself.

\begin{table}[t]
\centering
\caption{System Adaptation Score (1--5) per user trait for Neutral (N), Try (T) and Oracle (O), averaged over LLMs, domains and two judges; the last two rows show each judge. $^{***}$: $p<.001$ (Wilcoxon over 60 matched cells, Holm-corrected).}
\small
\setlength{\tabcolsep}{4pt}
\begin{tabular}{lccccc}
\toprule
\textbf{User trait} & \textbf{N} & \textbf{T} & \textbf{O} & \textbf{$\Delta$T$-$N} & \textbf{$\Delta$O$-$N} \\
\midrule
\textit{Extraversion}        & 2.02 & 3.46 & 4.74 & +1.44 & +2.72 \\
\textit{Agreeableness}       & 2.51 & 2.88 & 3.81 & +0.37 & +1.30 \\
\textit{Conscientiousness}   & 2.93 & 3.11 & 3.69 & +0.18 & +0.76 \\
\textit{Neuroticism}         & 2.65 & 3.82 & 4.17 & +1.17 & +1.52 \\
\textit{Openness}            & 1.84 & 2.77 & 4.20 & +0.93 & +2.36 \\
\midrule
\textit{Introversive}         & 2.18 & 2.27 & 2.53 & +0.09 & +0.35 \\
\textit{Disagreeable}        & 1.29 & 1.81 & 3.17 & +0.52 & +1.88 \\
\textit{Unconscientious}     & 1.88 & 2.04 & 2.65 & +0.16 & +0.77 \\
\textit{Stable}  & 2.71 & 2.71 & 2.76 & +0.00 & +0.05 \\
\textit{Closed}              & 2.20 & 2.09 & 2.31 & $-$0.11 & +0.11 \\
\midrule
\textbf{Average}    & \textbf{2.22} & \textbf{2.70} & \textbf{3.40} & \textbf{+0.48$^{***}$} & \textbf{+1.18$^{***}$} \\
\quad Gemma 4         & 1.38 & 2.06 & 2.86 & +0.68$^{***}$ & +1.49$^{***}$ \\
\quad GPT-5-mini    & 3.06 & 3.34 & 3.94 & +0.27$^{***}$ & +0.88$^{***}$ \\
\bottomrule
\end{tabular}
\label{tab:sys_adaptation}

\end{table}

\subsection{Human Evaluation}
\label{sec:human}

 We validated the pairwise satisfaction judges, our most subjective metric, through a human study on Prolific using Qualtrics. We sampled 175 dialogue pairs stratified across the three comparison types, ten traits, three LLMs, and both domains, with two annotators per pair (350 ratings; \$12/hour). Annotators chose which system the user would find more satisfying, or ``About the same,'' using the judges' criteria: helpfulness, tone, responsiveness, clarity, and fit to the user's communication style, excluding task success. Access condition and LLM identity were hidden, and dialogue order was randomized. We compare annotators with both LLM judges (Gemma 4 and GPT-5-mini) and report the average over the two judges.

Including ties, humans agreed with the LLM judges in 59.0\% of ratings on average ($\kappa{=}0.337$; Gemma 4 59.7\%, $\kappa{=}0.342$; GPT-5-mini 58.3\%, $\kappa{=}0.331$), above inter-annotator agreement (47.4\%, $\kappa{=}0.189$). Much of the remaining gap reflects calibration: the judges rarely call a tie (Gemma 4 0.6\%, GPT-5-mini 7.4\% of pairs), whereas annotators did so in 22\% of ratings. On the no-tie subset, agreement rises to 76.3\% on average ($\kappa{=}0.525$; Gemma 4 76.8\%, $\kappa{=}0.535$, $n{=}271$; GPT-5-mini 75.8\%, $\kappa{=}0.515$, $n{=}256$), against 65.1\% ($\kappa{=}0.304$) between annotators. Where the two judges agree with each other (74\% of pairs), agreement with humans on the no-tie subset reaches 83.2\% ($\kappa{=}0.663$). Only 7--8 of 175 pairs (${<}5\%$) had both annotators agreeing with each other against a judge, with no systematic pattern across conditions or traits. Both judges therefore track human observers when a clear winner exists, supporting the trends in \S\ref{sec:rq2}.

\section{Conclusion}
\label{sec:conclusion}
We studied whether LLMs can express personality in goal-directed task-oriented dialogue without weakening task completion, and whether adapting to the user's personality improves interaction quality. We used a controlled, training-free two-agent framework evaluated across three LLMs, two SGD domains, and the Big Five traits with their opposite-pole variants, and scored every metric with two independent LLM judges (Gemma 4 and GPT-5-mini).

LLM agents can express personality while maintaining strong task performance, but realization is uneven. Expressive traits such as \textit{Neuroticism} and \textit{Extraversion} are detected almost perfectly, whereas \textit{Introversive} and \textit{Closed} remain near-undetectable. This is consistent with behavioral absence carrying little lexical signal in short, schema-constrained dialogue. Personality adaptation raises pairwise user satisfaction under both adaptive conditions. Cue-responsive adaptation (\textit{Try}) also improves Constraint Satisfaction and Inform Rate, whereas explicit conditioning (\textit{Oracle}) does not. Both conditions reduce Truthfulness, revealing a tension between personalization and task grounding. The gains in task metrics and satisfaction are larger in the preference-driven Restaurant domain, where the system has more latitude to adapt.

How adaptation is implemented matters most. Explicit conditioning in \textit{Oracle} follows a label--behavior match effect: Oracle does not make the user's trait more visible, yet its gains in both condition-blind adaptation scores and satisfaction concentrate on traits whose behavior is already visible, as do judge-free elaboration counts. For \textit{Disagreeable} users, the label triggers strong tone mirroring in Qwen3 and Gemini, which judges rate as highly adapted but users would find less satisfying. Cue-responsive adaptation in \textit{Try} avoids this failure and is preferred over \textit{Oracle} overall. These findings are consistent across both judges, and the judges are reliable proxies for human observers: annotators agree with the judges more often than with each other, both including ties (59.0\% vs.\ 47.4\%; $\kappa{=}0.337$ vs.\ $0.189$) and on clear-winner cases (76.3\% vs.\ 65.1\%; $\kappa{=}0.525$ vs.\ $0.304$), rising to 83.2\% when the two judges agree. Overall, responding to observable behavior offers a more robust and practically deployable route to personality-aware task-oriented dialogue than conditioning on psychological labels. Validating this principle with real users and domains spanning a wider range of preference latitude is the key next step for future research.

\section*{Acknowledgment}
This work was supported in part by the National Science Foundation (NSF) under grant IIS-2401685.

\bibliographystyle{IEEEtran}

\bibliography{IEEEabrv}

\end{document}